\documentclass[lettersize,journal]{IEEEtran}
\usepackage{amsmath,amsfonts}
\usepackage{algorithmic}
\usepackage{algorithm}
\usepackage{array}
\usepackage[caption=false,font=normalsize,labelfont=sf,textfont=sf]{subfig}
\usepackage{textcomp}
\usepackage{url}
\usepackage{verbatim}
\usepackage{graphicx}
\usepackage{cite}

\usepackage{hyperref}

\usepackage{cleveref}
\usepackage{booktabs}
\usepackage{adjustbox}
\usepackage{balance}
\usepackage{placeins}

\newcommand{\mat}[1]{\mathbf{#1}}
\DeclareUnicodeCharacter{2265}{\ensuremath{\ge}}

\begin{document}

\title{Unraveling the Real Working Mechanism and Inherent Flaws of GAE: A Method for Interpreting Transformer Processes from an Economic Perspective}

\author{Yongjin Cui, Xiaohui Fan
\thanks{Yongjin Cui, Zhejiang University, cuiyongjin@zju.edu.cn; Xiaohui Fan, Zhejiang University, fanxh@zju.edu.cn}
}



\maketitle

\begin{abstract}
We observe a phenomenon that current algorithmic research in the field of explainable artificial intelligence primarily pursues better performance on several proxy metrics. On the one hand, these proxy metrics themselves are more or less flawed and cannot properly measure the quality of methods. On the other hand, metric-oriented research approaches often lead to the neglect of the rationality and interpretability of the methods themselves. Explainable artificial intelligence is abbreviated as XAI. The metric-driven research paradigm has resulted in a lack of interpretability of the relevant XAI methods themselves. Accordingly, there is a need for interpretability research on XAI methods, which can be playfully referred to as XXAI. This paper is one of our works on XXAI. This paper takes Generic Attention-model Explainability (GAE), a widely influential model interpretation method , or rather, XAI method that represents an important technical route, as the research object, and explores the real working mechanism and flaws of this method as well as the technical route it represents. Based on the conclusions of this study, it may be necessary to re-examine or verify GAE-related methods and their domain applications. We argue that GAE is an interpretation method that focuses on the attention process. After pointing out the working mechanism and flaws of GAE, we propose Cumulative Asset
Holdings (CAH), a more reasonable Transformer interpretation method integrating both process-based and feature-based ideas from an economic zero-sum games perspective. In addition, it is worth noting that our method is applicable to models with special tokens, where existing methods may suffer from limitations. The model simplification research method and the analysis of additive operations adopted in this study may provide inspiration for other research works in XAI.

\textit{This work has been submitted to the IEEE for possible publication. Copyright may be transferred without notice, after which this version may no longer be accessible.}

\end{abstract}
\section{Introduction}
Transformer was initially proposed and applied in the field of natural language processing \cite{1}, and has subsequently been successfully applied to computer vision \cite{6,7}, audio processing \cite{12,13}, multimodal learning \cite{21,22,23}, and other interdisciplinary fields such as pharmaceuticals \cite{28,29,30}. It has also spawned a series of large models \cite{38,41,42}, further advancing the development of AI. The proposal of Transformer marks a major turning point in the development of AI, and its remarkable performance has further motivated efforts to interpret its internal mechanisms.

Along with the remarkable progress of AI driven by Transformer, concerns have also emerged. These concerns mainly stem from the black-box nature of deep models, which may bring unknown issues, such as algorithmic safety in AI-based healthcare and autonomous driving, as well as algorithmic fairness across different groups.

It is worth noting that interpretability research differs substantially from traditional model training research. Model training has a clear objective: minimizing the loss. For example, in the training of image recognition models, both the training objective and evaluation metric are the recognition accuracy of the model, and the objective is reliable. Therefore, its performance evaluation method is well-defined, and model training is clearly metric-oriented. However, model interpretation often lacks well-defined evaluation metrics, and the few existing metrics are flawed proxy metrics.

Currently, there are two main approaches to evaluating the performance of model interpretation methods: qualitative and quantitative. Qualitative methods compare model interpretation results with human understanding of target features to verify their consistency. There are two types of quantitative methods. One is similar to qualitative evaluation, which compares the consistency between model interpretation results and feature regions in a batch manner. The other type perturbs the input based on model interpretation results—for example, removing corresponding elements in order of importance and observing the magnitude of changes in model outputs.

Consistency-based evaluation relies on a default assumption: human perception is the optimal solution, and the tested model is sufficiently good to achieve this optimal solution. Under this assumption, model interpretation results are compared with human understanding of features—the higher the consistency, the better the interpretation method. First, we cannot be fully certain that human cognition indeed represents the optimal solution; models may also learn other patterns on their own. Another issue is that we cannot guarantee that the tested model is sufficiently good. In this case, deviations in model interpretation results may stem from the interpretation method itself, insufficient model performance, or new patterns learned by the model that differ from human cognition.

Perturbation-based evaluation methods ignore the inherent regularities of data. The information conveyed by data depends on the relationships between elements. Quantitative evaluation methods based on perturbation essentially rely on marginal effects, which destroy the interdependence of data and introduce uncertainty, making it difficult to guarantee the accuracy and reliability of evaluation results. For example, the artifact phenomenon \cite{DBLP:conf/ijcai/Xie0CZ23} validates this issue; readers may refer to that paper for artifact cases. Regarding the limitations of marginal effects, we can illustrate them with a very simple everyday example. A man and a woman marry and have a child. If evaluated using marginal effects and assuming the absence of the man in this relationship, the child would not exist. Therefore, marginal-effect-based evaluation would attribute the entire contribution of having the child to the man, which is obviously incorrect.

Researchers in the field of XAI are often influenced by the mindset of model training research and still adopt a metric-oriented research approach, yet they overlook that the evaluation metrics for model interpretation results are unreliable proxy metrics. This metric-oriented research approach leads researchers to focus more on metric performance while neglecting the rationality and interpretability of the model interpretation methods themselves. This phenomenon perfectly exemplifies Goodhart's Law: when a measure becomes a target, it ceases to be a good measure. This is particularly true when the proxy metrics themselves are inherently flawed.

Model interpretation methods, or XAI methods, were originally developed to study the interpretability of models, but their own interpretability still needs to be investigated—herein referred to as XXAI. Existing studies have demonstrated that widely recognized results, such as Grad-ECLIP \cite{gradeclip}, TAM \cite{yuan2021explaining}, BT \cite{46}, DIX \cite{47}, and ODAM \cite{ODAM1,ODAM2}, have flaws or errors \cite{cui2026debunkinggradeclipcomprehensivestudy,cui2026IntegratedGradients,cui2026neglectedbaselinemodelinterpretation}.

Among numerous Transformer interpretation methods, GAE \cite{48} holds a very important position. It formally provides a highly representative technical pipeline: first assigning weights to attention maps (GAE uses the positive gradient of attention maps as weights), and then performing attention rollout\cite{58} across all layers to obtain the final interpretation scores:
\begin{align}
\bar{A}^{(l)} &= I + \mathbb{E}_H\!\big((\nabla A^{(l)})^+ \odot A^{(l)}\big), \label{eq:gae1}\\
\mathrm{rollout} &= \bar{A}^{(L)} \cdot \bar{A}^{(L-1)} \cdots \bar{A}^{(2)} \cdot \bar{A}^{(1)}, \label{eq:gae2}\\
C &= \mathrm{rollout}[0, 1:]. \label{eq:gae3}
\end{align}
where $A^{(l)}$ is the attention map of the $l$-th layer, $\mathbb{E}_H$ denotes the expectation over multiple heads, $\odot$ denotes the Hadamard product, $(\cdot)^+$ takes the positive part, and $C$ is the overall contribution originating from the cls token.

\textbf{Note that the original GAE paper introduces this computation pipeline from the perspective of correlation propagation, while the actual calculation process is completely equivalent.} 

\begin{figure*}[!t]
  \centering
  \includegraphics[width=1.5\columnwidth]{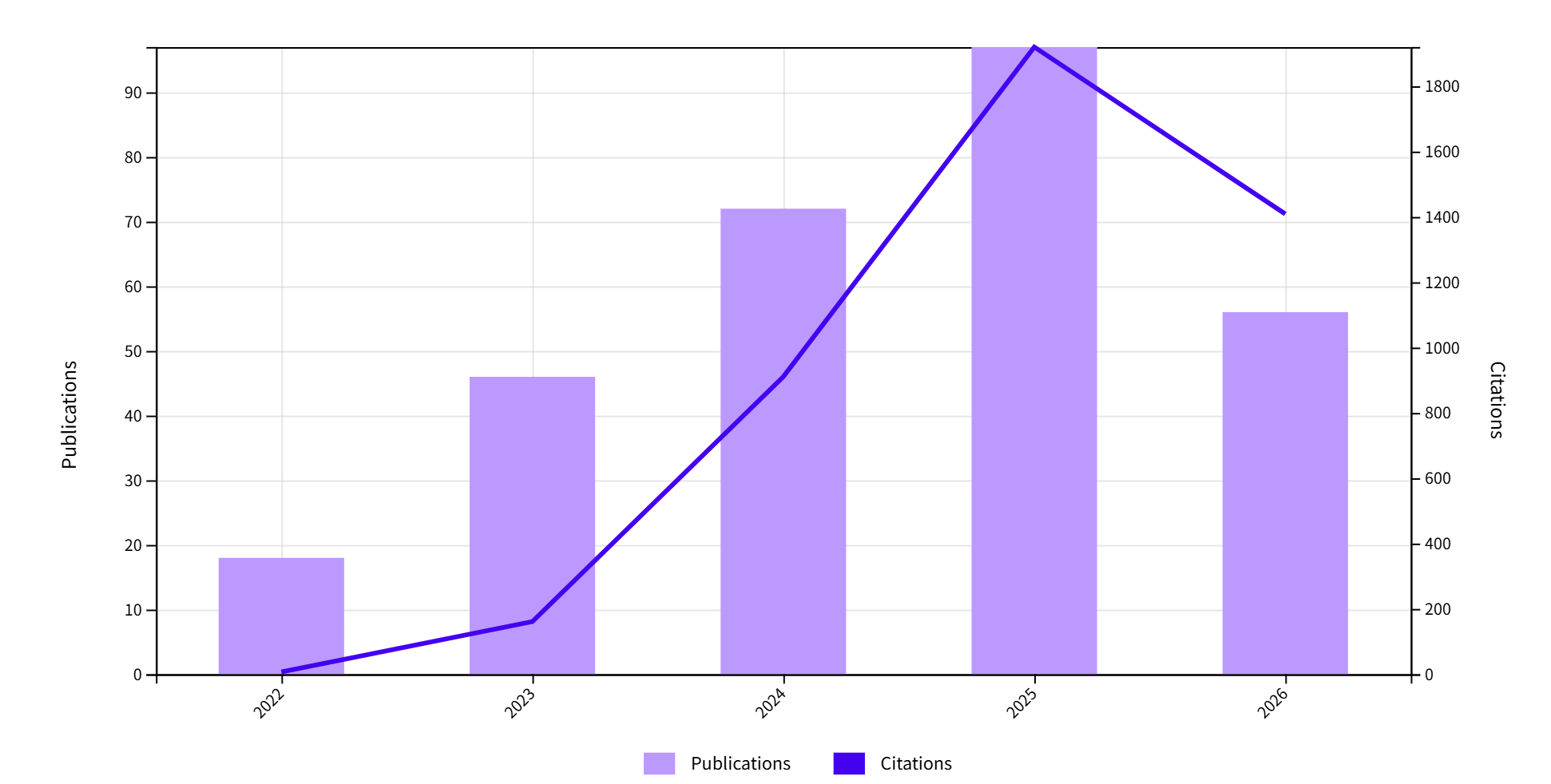}
  \caption{Citation statistics of GAE from Web of Science as of August 11, 2026}
  \label{gaetongji}
  \end{figure*}

However, the working mechanism of this technical pipeline has not been fully and effectively studied, yet it has been widely accepted due to its outstanding performance on existing flawed proxy metrics. As of August 11, 2026, statistics from Web of Science show that GAE has been cited 4,411 times across 289 publications (\Cref{gaetongji}).

GAE has been applied in many fields and has influenced numerous subsequent methods; see \Cref{relatedwork} for details. Therefore, it is highly necessary to conduct a detailed and in-depth investigation of the working mechanism of GAE.

This paper conducts an in-depth discussion on the working mechanism of GAE, and clarifies its real mechanism and inherent flaws. \textbf{Based on the conclusions of this study, it may be necessary to re-examine or verify GAE-related methods and their domain applications.}

GAE is essentially a gradient-based method built on linear approximations of numerous nonlinear operations in the model, which introduces certain approximation errors and fails to satisfy Sensitivity \cite{61}. Integrated Gradients is a more accurate model interpretation method, but it cannot be applied to attention maps. Since attention maps are generated from features, the two are not independent of each other, making it impossible to perform Integrated Gradients operations on attention maps alone; see the mechanism of Integrated Gradients for details \cite{61,cui2026neglectedbaselinemodelinterpretation}. In addition, in Transformers with special tokens—such as ViT \cite{6}, which contains the cls token—applying Integrated Gradients assigns part of the output contribution to the cls token. Except for the initial cls token, the cls tokens in other layers all contain contributions from other patch tokens. Accordingly, it is difficult to further allocate the contribution of the cls token to individual patch tokens.

That is to say, Integrated Gradients is more accurate, but it cannot be applied to attention maps (the process-based perspective), nor can it be easily applied to interpreting Transformers with special tokens (the feature-based perspective). Having criticized GAE to a certain extent, we ought to propose a new interpretation method. Finally, we integrate process-based and feature-based perspectives, and propose a Transformer interpretation method from the perspective of zero-sum games in economics.

\section{Related Work}\label{relatedwork}
\subsection{Generic Attention-model Explainability (GAE)}
\subsubsection{GAE and Its Applications}
Transformer interpretability has long been one of the core challenges in deep learning research. Among various methods, GAE approach proposed by Chefer et al. \cite{48} is a milestone as a foundational framework for interpreting Transformers. The core idea of GAE is to compute a modified attention matrix $\bar{A} = \mathbb{E}_h[(\nabla A \odot A)^+]$, where $\odot$ denotes the Hadamard product, $\mathbb{E}_h$ averages over attention heads and discards negative contributions. Relevance is then propagated across layers via the rollout \cite{58} update rule $R \leftarrow R + \bar{A}R$.

GAE has been widely applied across numerous fields far beyond its original vision-language context. In the medical field, the TRUSformer model applies Chefer et al.'s technique to determine the contribution of each region of interest (ROI) in micro-ultrasound sequences to the prediction of ``benign" or ``malignant" biopsy core categories. A relevance map $R$ initialized as an identity matrix is propagated through attention layers with the update rule $R = R + \bar{A}R$ \cite{WOS:000993658000001}. The ECOLE model uses GAE to visualize which read-depth signal segments are considered important when calling copy number variations in whole-exome sequencing data \cite{WOS:001142794000883}. In audio processing, GAE has been adapted to speech foundation models to reveal acoustic features associated with depression detection based on gradient-weighted attention maps \cite{WOS:001613931400476}; and through the Gradient Average Transformer Relevancy (GATR) method, which performs gradient-weighted averaging of output token relevance to explain Transformer-based audio deepfake detectors \cite{WOS:001548470300011}. In agriculture, a multimodal crop yield simulation model employs attention rollout and weighted modality activation for feature and modality attribution \cite{WOS:001597510200001}. The GAE framework has also been extended to the iTransformer time-series forecasting architecture, adapting Chefer's interpretation method to identify key variables and their correlations \cite{WOS:001706337700677}. Other notable applications include handwritten text image editing—which uses Chefer et al.'s ATT method combined with an OCR model for text localization \cite{WOS:001721216200029}—and medical image retrieval, where Chefer et al.'s attention rollout is implemented to visualize the spectrogram regions driving model decisions \cite{WOS:001482498600092}. The breadth of these applications demonstrates that GAE has become a universal Transformer interpretation tool across medical imaging, genomics, audio processing, agriculture, time-series analysis, and multimodal learning.

\subsubsection{GAE as a Foundation for Subsequent Method Development}
Beyond direct applications, GAE has had a profound impact on the development of subsequent interpretation methods. Many methods either directly incorporate GAE or adopt its core design principle: \emph{weighting attention maps (not limited to gradients), followed by cross-layer attention rollout}.

\paragraph{Methods That Directly Incorporate GAE}
Several methods are explicitly built upon GAE, using its gradient-weighted attention formula as a core component. The Gradient Rollout (GR) technique proposed in Iterated Integrated Attributions (IIA) \cite{WOS:001159644302032}, as well as the GR adopted in Deep Integrated Explanations (DIX) \cite{47}, is explicitly defined as ``a modified version of the Attention Rollout (AR) method, differing in that the computation includes the Hadamard product of each attention map with its gradient, rather than relying solely on the attention map itself". It takes the form $A'_b = I + E_h(A_b \circ G_b)$, followed by $GR = A'_{1} \cdot A'_{2} \cdots A'_{B}$ \cite{WOS:001159644302032}. The Mixture of Baseline Distributions (MBD) method also positions GAE as the state-of-the-art interpretation baseline for ViT that it benchmarks against \cite{WOS:001349579600058}. The EATI method explicitly states: ``following Chefer et al.". It computes gradient-guided averaging across multiple self-attention heads to obtain the final self-attention map $\tilde{A} = \mathbb{E}_h[(\nabla A \odot \hat{A})^+]$, and then ``following Rollout", integrates the attention map with the relevance matrix to propagate per-feature relevance via the update rule $R^{(l)} = R^{(l-1)} + \tilde{A}^{(l)} R^{(l-1)}$ \cite{WOS:001826339200001}. The Alert-ME defense framework similarly computes attention maps as the Hadamard product of attention weights and gradients, averaged across all heads after excluding negative contributions \cite{WOS:001714495200046}. Although the GradToken method introduces a gradient decoupling mechanism, it also adopts a cross-layer attention aggregation strategy: it first aggregates low-level attention weight matrices using the Rollout method, and then propagates class relevance via element-wise multiplication \cite{WOS:001358014000001}. The MGA-CLIP framework aligns with classic attribution paradigms such as Grad-CAM and attention Rollout, performing gradient attribution at the last Transformer layer and weighting the value features of the last layer with accumulated gradient evidence \cite{WOS:001779926800001}. A systematic review of Transformer interpretability \cite{WOS:001210760200001} further describes Attention Gradient (AGrad) and Relevance Propagation with Attention Gradient (RePAGrad) as computing relevance scores for each token via the product of loss gradient and attention weights $R_j = \sum_i [R'_{i,j} \cdot (-\frac{\partial L}{\partial A_j} \cdot A_j)]$, followed by recursive LRP backpropagation—directly echoing GAE's gradient-weighted attention scheme \cite{WOS:001210760200001}.

\paragraph{Methods Adopting the Core Design Principle}
Beyond direct adoption, a broader class of methods follows the conceptual blueprint established by GAE—weighting attention maps (not limited to gradients) followed by cross-layer attention propagation—while introducing alternative weighting mechanisms. The MATEX framework combines gradient signals with layer-wise attention rollout and clinical spatial cues to generate anatomically grounded explanations for medical vision-language models, and extends attention rollout to fuse spatial features across Transformer layers \cite{WOS:001760782300014}. The SkipPLUS method uses Rollout as a general aggregation operator for attribution maps, updating the cumulative map as $R_{i+1} = (0.5I + 0.5\bar{T}_i)R_i$, with normalization that ``mimics the behavior of attention weights" \cite{WOS:001327781700021}. The Thoraxformer model extracts lesion-aware attention tokens by averaging attention maps across heads followed by renormalization \cite{WOS:001694266300001}. The HIT-EC enzyme prediction model combines attention flow with gradient-based relevance propagation to provide a comprehensive perspective on the model's decision-making process \cite{WOS:001676670100004}. The Independent Block-level Attribution (IBA) method distributes relevance within class-semantic blocks via Block-level Layer-wise Relevance Propagation (BLRP) and Block-level Gradient (BGrad) \cite{WOS:001811609400007}. The RAM (Relevance Attention Map) framework proposes a unified framework that shares the same interpretation mechanism for both Transformer and CNN models \cite{WOS:001811609400007}. LRP modifications for VisualBERT adapt Chefer et al.'s scheme, allowing attention to attribute in a nonlinear manner by combining gradient and relevance information across layers \cite{WOS:000982335600007}.

In summary, GAE established a paradigm that can be summarized as follows: \emph{first weight attention maps (usually via element-wise product with their gradients or other relevance signals), then propagate the weighted maps across layers via attention rollout}. This two-stage ``weight-then-rollout" design has been adopted, extended, and re-implemented by numerous subsequent works, establishing GAE as one of the most influential interpretation methods for Transformer architectures.

\subsection{Integrated Gradients (IG)}
The emergence of IG addresses the issue that gradients fail to satisfy Sensitivity\cite{61}, and IG satisfy Implementation Invariance\cite{61} at the same time.
  
The core principle of IG is to focus on every step of the input contribution's change from the input baseline to the current input, rather than merely concentrating on the final state (gradient-based method such as GAE).
IG is defined as the path intergral of the gradients along the straightline path from the baseline $x' \in R^{n}$ to the input $x \in R^{n}$. The integrated gradient along the $i_{th}$ dimension is defined as follows.
\begin{equation}\label{eq:IG}
  \begin{aligned}
      &\text{IG}_i(x) \\
      & ::=F(x)-F(x')\\
      & ::=\int_{\alpha=0}^1\frac{\partial F(x^{\prime}+\alpha(x-x^{\prime}))}{\partial (x_i^{\prime}+\alpha(x_i-x_i^{\prime}))}  \frac {\partial (x_i^{\prime}+\alpha(x_i-x_i^{\prime}))}{\partial \alpha} d\alpha\\
      &=(x_i-x_i^{\prime})\int_{\alpha=0}^1\frac{\partial F(x^{\prime}+\alpha(x-x^{\prime}))}{\partial (x_i^{\prime}+\alpha(x_i-x_i^{\prime}))}d\alpha
  \end{aligned}
  \end{equation}
The calculation process adopts a straightline path from the baseline input to the current input; for the advantages of this approach, please refer to the original paper.
The integral of IG can be efficiently approximated via a summation (Riemann Sum). 
\begin{equation} \label{eq:IG1}
  \begin{aligned}
    &\text{IG}_i^{approx}(x)\\
    &::=\frac{(x_i-x_i^{\prime})}{m}\times\Sigma_{k=1}^m\frac{\partial F(x^{\prime}+\frac{k}{m}(x-x^{\prime}))}{\partial (x_i^{\prime}+\frac{k}{m}(x_i-x_i^{\prime}))}
  \end{aligned}
\end{equation}
Here $m$ is the number of steps in the Riemann approximation of the integral. 
In \Cref{eq:IG1}, we can find that the gradient method is actually a special case of IG when $m = 1$, and the baseline $x^{\prime}$ is all-zero.
Formally, it simply performs a simple linear approximation of a complex nonlinear operation, as show in \Cref{eq:G}.
\begin{equation} \label{eq:G}
  \begin{aligned}
    &\text{G}_i(x)\\
    &::=x_i\times\frac{\partial F(x)}{\partial (x_i)}
  \end{aligned}
\end{equation}
Cui et al. proposed a more rigorous improvement to IG, introducing a scheme that determines the baseline of the entire model from the input end. All interpretations of model interpretation results are based on this baseline, and the quality of interpretation results is evaluated based on attribution error rather than flawed proxy metrics; see the original paper for details \cite{cui2026neglectedbaselinemodelinterpretation}. All content related to IG in this paper is based on this improved version of IG.

\section{Unraveling the Real Working Mechanism and Inherent Flaws of GAE}
\subsection{Research Rationale}
Attention rollout is essentially a simplification of the model: it only retains the attention process of the original model and represents residual connections with identity matrices, while ignoring other model processes. GAE adds a weight to attention maps—the positive gradient of the attention map with respect to the output to be explained—and then performs attention rollout. Therefore, we can investigate the working mechanism of GAE from the perspective of model simplification.

\subsection{Model Simplification and Initialization}
\begin{figure}[!t]
  \centering
  \includegraphics[width=0.15\columnwidth]{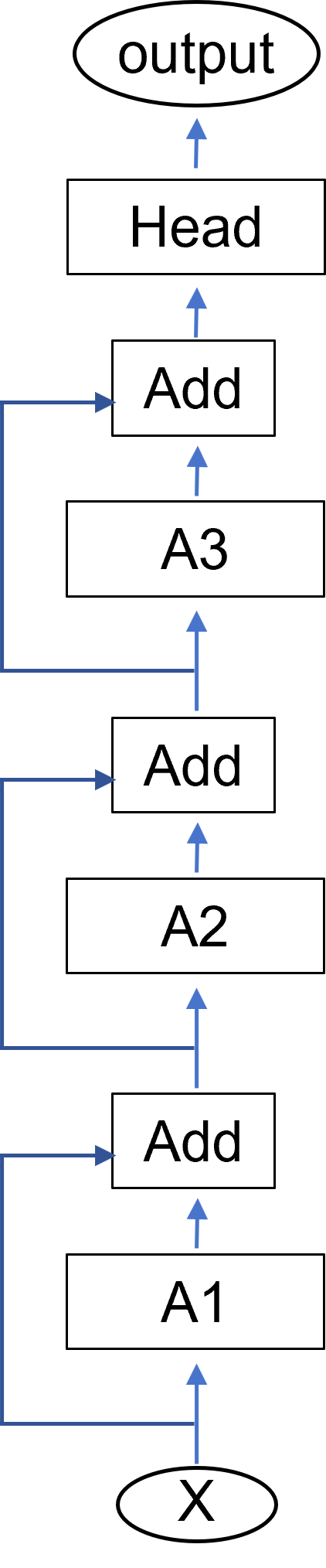}
  \caption{Simplification of ViT.}
  \label{jianhuamoxing}
  \end{figure}

We take ViT \cite{6} as an example for simplification (\Cref{jianhuamoxing}), considering only inputs, three layers of single-head attention, residual connections, a head and a single output, with specific initializations.

The input matrix is $\mat{X} \in \mathbb{R}^{3 \times 2}$, with the first row being the cls token. The three single-head attention maps are $\mat{A}_1, \mat{A}_2, \mat{A}_3 \in \mathbb{R}^{3 \times 3}$. The specific initialization values are as follows (note that any other initialization is acceptable, and the experimental conclusions are not limited to the initialization provided in this paper):
\begin{align} \label{eq:x1}
\mat{X} &= \begin{pmatrix} 0.0 & 0.0 \\ 0.01 & 0.02 \\ 0.03 & 0.01 \end{pmatrix}, \\[6pt]
\mat{A}_1 &= \begin{pmatrix} 0.8 & 0.1 & 0.1 \\ 0.7 & 0.1 & 0.2 \\ 0.6 & 0.2 & 0.2 \end{pmatrix}, \\[6pt]
\mat{A}_2 &= \begin{pmatrix} 0.4 & 0.3 & 0.3 \\ 0.3 & 0.6 & 0.1 \\ 0.5 & 0.3 & 0.2 \end{pmatrix}, \\[6pt]
\mat{A}_3 &= \begin{pmatrix} 0.5 & 0.3 & 0.2 \\ 0.4 & 0.3 & 0.3 \\ 0.3 & 0.4 & 0.3 \end{pmatrix}.
\end{align}

Residual connections are represented as $\mat{I}_3$, a $3 \times 3$ identity matrix. We then define the residual-fused attention map:
\begin{equation}
\widetilde{\mat{A}}_k = \mat{A}_k + \mat{I}_3, \quad k = 1, 2, 3,
\end{equation}
The forward propagation process is defined as:
\begin{equation}
\mat{Z}_1 = \widetilde{\mat{A}}_1 \mat{X}, \quad
\mat{Z}_2 = \widetilde{\mat{A}}_2 \mat{Z}_1, \quad
\mat{Z}_3 = \widetilde{\mat{A}}_3 \mat{Z}_2.
\end{equation}
The model output is the sum of elements in the cls token of the third-layer attention output:
\begin{equation}
\mathcal{L} = \sum_{j=0}^{1} [\mat{Z}_3]_{0j}.
\end{equation}

\subsection{Attention-Map-Based Interpretation Process}
Following the idea of GAE, we set the loss $\mathcal{L}$ as the output, and use backpropagation to compute the gradient of each residual-fused attention map $\frac{\partial \mathcal{L}}{\partial \widetilde{\mat{A}}_k}$.

On this basis, we construct Hadamard product matrices, i.e., performing gradient weighting using gradients as weights.
\textbf{We first consider the case of full gradients.}

\paragraph{Weighting of residual-fused attention maps}
\begin{equation}
\widetilde{\mat{\mathbb{A}}}_k = \widetilde{\mat{A}}_k \odot \frac{\partial \mathcal{L}}{\partial \widetilde{\mat{A}}_k}, \quad k = 1, 2, 3.
\end{equation}
Define $sum{\widetilde{\mat{\mathbb{A}}}}_k$ as the sum of all elements in $\widetilde{\mat{\mathbb{A}}}_k$:
\begin{equation}
sum\mat{\widetilde{\mat{\mathbb{A}}}}_k = \sum \widetilde{\mat{\mathbb{A}}}_k
\end{equation}
In this model case, the loss used for gradient computation is
\begin{equation}
\mathcal{L}=output=0.0955
\end{equation}
And
\begin{equation}\label{eq:sumak}
sum\mat{\widetilde{\mat{\mathbb{A}}}}_1=sum\mat{\widetilde{\mat{\mathbb{A}}}}_2=sum\mat{\widetilde{\mat{\mathbb{A}}}}_3=\mathcal{L}=output=0.0955
\end{equation}

This is because this simplified approximate model is completely linear with $bias=0$ (passing through the origin). The gradient weighting of residual-fused attention maps essentially represents the contribution of the residual-fused attention map to the output, and the sum of all contributions equals the output. All gradient-weighted model interpretation methods in the field of XAI are based on this approximate assumption: they simplify and approximate complex nonlinear processes as linear processes (including discarding some operations—for example, when applied to ViT, this is equivalent to discarding layer norm before attention operations, or treating features before and after layer norm as equivalent), and assume that the linear approximation process passes through the origin, i.e., $bias=0$. These two assumptions are the fundamental source of error for all gradient-weighted interpretation methods.

The above analysis mainly clarifies a core idea: gradient-weighted attention represents the contribution of each attention unit to the model output, and the sum of their values equals the model output.

After gradient weighting, we proceed to perform attention rollout. It is particularly important to note that GAE does not compute gradient weighting for the residual connection $I$, and only performs gradient weighting on the original attention maps. Therefore, we next perform calculations following GAE's gradient weighting scheme.

\paragraph{Weighting of attention maps}
\begin{equation}
{\mat{\mathbb{A}}}_k = {\mat{A}}_k \odot \frac{\partial \mathcal{L}}{\partial {\mat{A}}_k}, \quad k = 1, 2, 3.
\end{equation}
Simple derivation shows that $\frac{\partial \mathcal{L}}{\partial {\mat{A}}_k}=\frac{\partial \mathcal{L}}{\partial \widetilde{\mat{A}}_k}$. The subsequent attention rollout operation is shown in \Cref{eq:zhankai}:
\begin{equation}\label{eq:zhankai}
  \begin{aligned}
rollout=
&(\mathbb{A}_3+I_3)(\mathbb{A}_2+I_3)(\mathbb{A}_1+I_3)\\
=
&\mathbb{A}_3\mathbb{A}_2\mathbb{A}_1\\
&+\mathbb{A}_3\mathbb{A}_2
+\mathbb{A}_3\mathbb{A}_1
+\mathbb{A}_2\mathbb{A}_1\\
&+\mathbb{A}_3
+\mathbb{A}_2
+\mathbb{A}_1\\
&+I_3
\end{aligned}
\end{equation}

There are four components in \Cref{eq:zhankai}: triple matrix multiplication terms, double matrix multiplication terms, first-order terms, and the identity matrix term.
\textbf{Among them, since GAE only takes the scores of corresponding patch tokens from the cls token, the identity matrix term $I_3$ has no effect on the final result.}

We focus here on the rationality of repeated matrix multiplications.
\Cref{eq:zhankai} involves repeated matrix multiplication of gradient-weighted attention maps. As we explained in relation to \Cref{eq:sumak}, gradient weighting actually represents the contribution of each element to the output; they are no longer plain attention maps that can be repeatedly multiplied as attention weights in attention rollout. Put more plainly, the weighting operation endows attention maps with physical meaning, and they are no longer merely weights that can be operated on interchangeably.

Here is a simple example: suppose a similar model is designed to predict the length $L$ of an observed object from input data. In this case, all gradient-weighted attention values have clear physical meanings, or specific units such as meters ($m$), indicating that attention $A_{ij}$ at a certain position contributes $L_{ij}$ to the output $L$. It is a common knowledge in numerical computation that addition and subtraction require the attributes or dimensions of the operands to be consistent.
For example, the output here is length prediction with dimension $m$. Double matrix multiplication terms represent area with unit $m^2$, and triple matrix multiplication terms represent volume with unit $m^3$. Adding values of three different dimensions—length, area, and volume—has no practical physical meaning.

Then why does GAE still achieve reasonable performance? The performance here refers to consistency with human cognition in visual qualitative judgment, ability to indicate target feature regions, and rationality in line with human perception in perturbation experiments. Taking the ViT\_base model as an example, this is because the model output is typically very small in magnitude. If the probability output is taken as the loss, the loss is less than 1, and the value distributed across different attention units is generally far less than 1. If the logit output is taken as the loss, the logit output is generally less than 20. Distributed across $197*197$ attention units, the weighted attention values are also generally far less than 1. Note that GAE discards negatively weighted attention values. Even if they were retained, the absolute values of weighted attention would generally still be far less than 1.

This causes higher-order (second-order and above) matrix multiplication terms to be much smaller in magnitude than the weighted attention values. The dominant contribution to the final result comes from the accumulation of weighted attention across layers, which we refer to as the accumulation of first-order terms.

In our simplified model,
\begin{equation}\label{eq:20}
  \mathbb{A}_3+\mathbb{A}_2+\mathbb{A}_1 = \begin{pmatrix}
      0.0234 & 0.0472 & 0.0496 \\
      0.0006 & 0.0104 & 0.0095 \\
      0.0007 & 0.0068 & 0.0079
    \end{pmatrix}
  \end{equation}
  \begin{align}\label{eq:21}
    &\mathbb{A}_3\mathbb{A}_2\mathbb{A}_1
    +\mathbb{A}_2\mathbb{A}_1
    +\mathbb{A}_3\mathbb{A}_1
    +\mathbb{A}_3\mathbb{A}_2 \notag \\
    &= \begin{pmatrix}
  1.0562\text{e-}4 & 1.0145\text{e-}3 & 1.3107\text{e-}3 \\
  0 & 3.3259\text{e-}5 & 7.3551\text{e-}5 \\
  0 & 2.1446\text{e-}5 & 3.8147\text{e-}5
    \end{pmatrix}
    \end{align}

In the above initialized model, the conclusion is consistent with our hypothesis: $\mathbb{A}_3+\mathbb{A}_2+\mathbb{A}_1$ dominates. However, there also exist cases where dimensionally meaningless terms $\mathbb{A}_3\mathbb{A}_2\mathbb{A}_1
+\mathbb{A}_3\mathbb{A}_2
+\mathbb{A}_3\mathbb{A}_1
+\mathbb{A}_2\mathbb{A}_1$ dominate. For example, if we set
\begin{align}
  \mat{X} &= \begin{pmatrix} 0.0 & 0.0 \\ 100 & 200 \\ 300 & 100 \end{pmatrix}
  \end{align}
then the dominant terms become $\mathbb{A}_3\mathbb{A}_2\mathbb{A}_1
+\mathbb{A}_3\mathbb{A}_2
+\mathbb{A}_3\mathbb{A}_1
+\mathbb{A}_2\mathbb{A}_1$(\Cref{duocixiang}), rather than $\mathbb{A}_3+\mathbb{A}_2+\mathbb{A}_1$(\Cref{dancixiang}).
\begin{equation}\label{dancixiang}
  \mathbb{A}_3+\mathbb{A}_2+\mathbb{A}_1 = \begin{pmatrix}
    233.5 & 472.5 & 495.8 \\
    6.3 & 103.5 & 95.4 \\
    7.0 & 67.8 & 79.2
    \end{pmatrix}
  \end{equation}
\begin{align}\label{duocixiang}
  &\mathbb{A}_3\mathbb{A}_2\mathbb{A}_1
  +\mathbb{A}_2\mathbb{A}_1
  +\mathbb{A}_3\mathbb{A}_1
  +\mathbb{A}_3\mathbb{A}_2 \notag \\
  &= \begin{pmatrix}
    1.0553\text{e+}4 & 4.7880\text{e+}6 & 8.5766\text{e+}6 \\
    0 & 3.3245\text{e+}3 & 7.3552\text{e+}3 \\
    0 & 2.1446\text{e+}3 & 3.8337\text{e+}3
  \end{pmatrix}
  \end{align}

It should be particularly emphasized that what we present here is one possibility, so we do not discuss changes in attention maps when inputs vary.

The reason GAE works well in many models is that many models conform to the scenario where $\mathbb{A}_3+\mathbb{A}_2+\mathbb{A}_1$ dominates.

We verify this using ViT as an example. We uniformly sample 5,000 images from ILSVRC-2012 ImageNet \cite{62} for experiments, and set the explanation class as the image label. We statistically analyze the numerical distribution of gradient-weighted average attention maps ($\mathbb{E}_H\!\big((\nabla A^{(l)})^+ \odot A^{(l)}\big)$ in \Cref{eq:gae1}).
\begin{table}[!t]
  \caption{ViT\_base Model Experiment: Distribution of 2,328,540,000 Gradient-Weighted Average  Attention Values}\label{tab:tongji1}
  \resizebox{\linewidth}{!}{%
  \begin{tabular}{@{}lllllll@{}}
  \toprule
  Mean       &  Std & \textgreater{}=1 & \textgreater{}=0.5 & \textgreater{}=0.1 & \textgreater{}=0.01 & \textgreater{}=0.001 \\ \midrule
  0.00002120 & 0.00027230                 & 4                & 36                 & 1066               & 143113              & 4718456              \\ \bottomrule
  \end{tabular}
  }
\end{table}

As shown in \Cref{tab:tongji1}, the mean value of the weighted average attention map distribution is 0.00002120 with a standard deviation of 0.00027230, making the contribution of higher-order terms almost negligible.

\begin{figure}[!t]
  \centering
  \includegraphics[width=0.5\columnwidth]{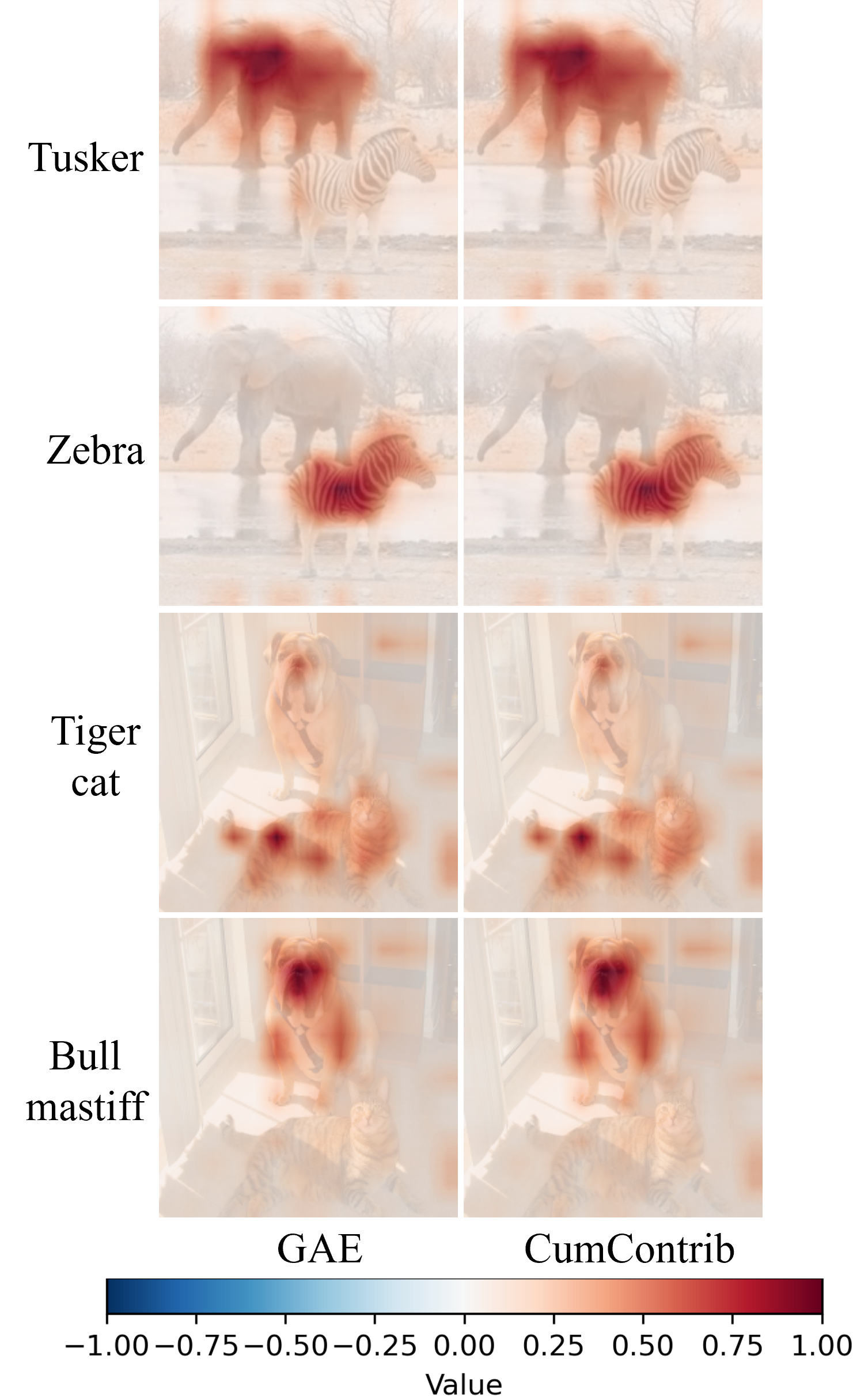}
  \caption{ViT\_base interpretation cases.(CumContrib: Cumulative Contribution of the first-order terms.)}
  \label{7_zhiguanduibi}
  \end{figure}

We provide several intuitive demonstration cases, as shown in \Cref{7_zhiguanduibi}.

\begin{figure}[!t]
  \centering
  \includegraphics[width=1\columnwidth]{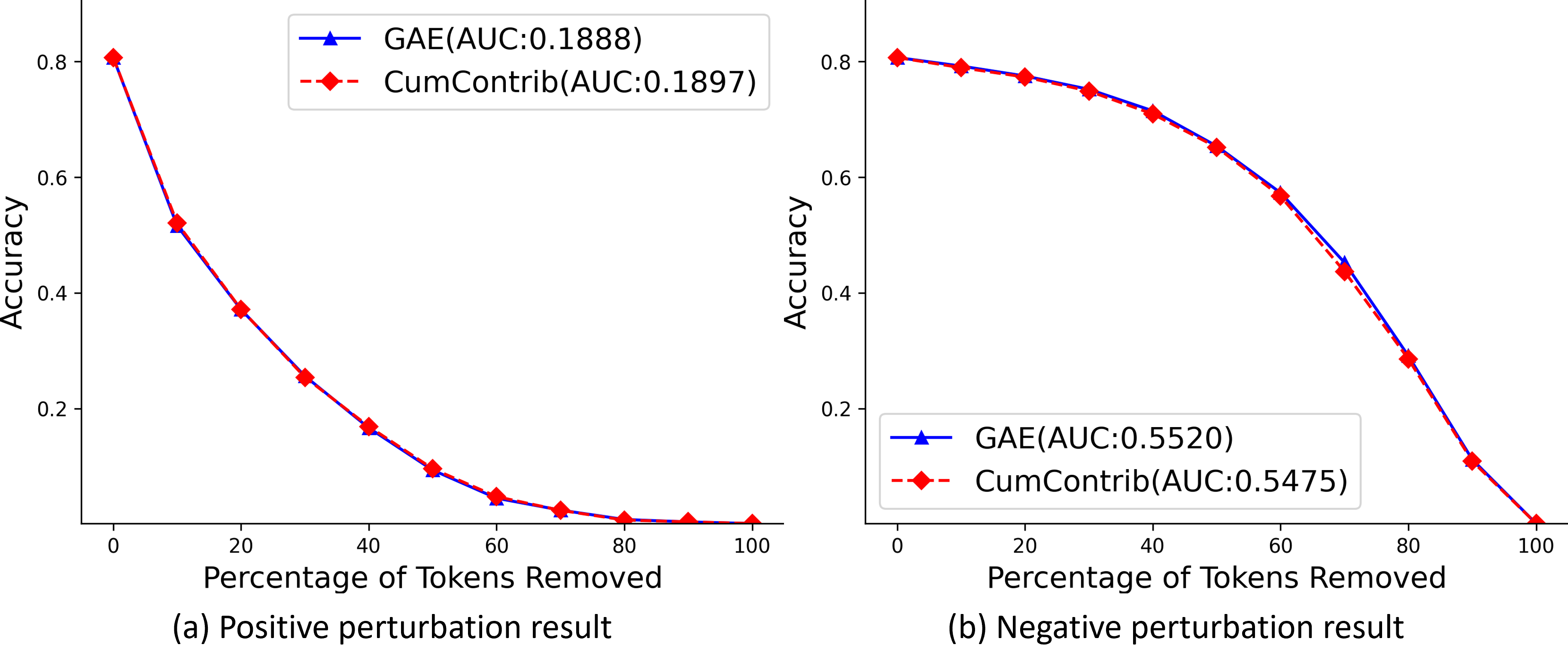}
  \caption{Perturbation Experiment Results.(CumContrib: Cumulative Contribution of the first-order terms.)}
  \label{7_raodong}
  \end{figure}

We conduct perturbation experiments, performing token removal perturbations based on the interpretation results. The experimental results are shown in \Cref{7_raodong}: the results of GAE are nearly identical to those of the first-order accumulation experiment (Cumulative Contribution:CumContrib).
Note that the purpose of our experiments here is to verify that the two are sufficiently close, not to compare their performance. Nor do we fully endorse this experimental evaluation method. This evaluation method is essentially based on marginal effects, which destroys the correlations between features and is inherently flawed.

This issue of GAE has not been exposed because the values of weighted average attention are generally very small.
We supplement the data distributions of weighted average attention maps for the LXMERT \cite{248} and DETR \cite{225} models following the experimental design of the original GAE paper. Both experiments use the same models and datasets as in the original GAE paper.

A pre-trained LXMERT is employed to extract weighted average attention for a randomly selected subset of 10,000 samples from the validation set of the VQA dataset \cite{251}. The results are shown in \Cref{tab:tongji2}.
\begin{table}[!t]
  \caption{LXMERT Model Experiment: Distribution of 160,366,014 Gradient-Weighted Average Attention Values}\label{tab:tongji2}
  \resizebox{\linewidth}{!}{%
  \begin{tabular}{@{}lllllll@{}}
  \toprule
  Mean       &  Std & \textgreater{}=1 & \textgreater{}=0.5 & \textgreater{}=0.1 & \textgreater{}=0.01 & \textgreater{}=0.001 \\ \midrule
  0.00018353508 & 0.00837882             & 0               & 0                & 661             & 133872              & 4394373              \\ \bottomrule
  \end{tabular}
  }
\end{table}

For the DETR experiment, we use 5,000 samples of the MSCOCO \cite{coco} validation set. We first filter the queries to retain only those with a classification probability exceeding 50\% for class-specific explanation.  
Different from the original GAE experiments, to reduce computational cost, we compute the weighted average attention solely based on the maximum class logit corresponding to the first qualified bounding box in each sample.
The statistical results are shown in \Cref{tab:tongji3}.
\begin{table}[!t]
  \caption{DETR Model Experiment: Distribution of 26,530,171,134 Gradient-Weighted Average Attention Values}\label{tab:tongji3}
  \resizebox{\linewidth}{!}{%
  \begin{tabular}{@{}lllllll@{}}
  \toprule
  Mean       &  Std & \textgreater{}=1 & \textgreater{}=0.5 & \textgreater{}=0.1 & \textgreater{}=0.01 & \textgreater{}=0.001 \\ \midrule
  0.000003 & 0.001606                 & 1025                & 2546                &  23668               & 468043              & 6521921              \\ \bottomrule
  \end{tabular}
  }
\end{table}

In summary, GAE's interpretation of the model is essentially the accumulation of first-order terms: it sums the contribution of each layer's attention to the output to measure the contribution of a specific patch to the output throughout the entire attention process. The remaining higher-order matrix multiplications are, on the one hand, theoretically unreasonable, and on the other hand, their magnitudes are too small to have a noticeable impact on the final result. However, there is no guarantee that higher-order terms will not cause severe interference in some special models. It may be necessary to re-examine or verify GAE-related methods and their domain applications. Meanwhile, the error of GAE also stems from the linear approximation of complex nonlinear processes.

\section{A Method for Interpreting Transformer Processes from an Economic Perspective}
Having demonstrated the working mechanism and flaws of GAE above, we here emphasize two other important properties of GAE: gradient-based and process-based. GAE is essentially a gradient-based method. Gradient-based methods crudely approximate complex nonlinear processes as linear ones, and are essentially single-step IG. Improving upon GAE using IG may seem like a feasible approach. However, attention maps are not actually independent of Transformer's Value (V) features, which prevents us from directly applying IG on attention maps. This point requires a more detailed understanding of the computational principle of IG \cite{61,cui2026neglectedbaselinemodelinterpretation}.

We therefore explore whether we can directly adopt Cui et al.'s IG scheme \cite{cui2026neglectedbaselinemodelinterpretation} for model interpretation. However, in Transformers with special tokens—such as ViT \cite{6}, which contains the cls token—applying IG assigns part of the output contribution to the cls token. Except for the initial cls token, the cls tokens in other layers all contain contributions from other patch tokens, making it difficult to further allocate the cls token's contribution to individual patch tokens. In this case, model interpretation based solely on features from a single layer can hardly attribute the output to all patch tokens. Under such circumstances, the process-based perspective may offer a promising technical route.

Based on the above two points, we start to explore integrating process-based and feature-based model interpretation ideas, and conduct Transformer interpretation from the perspective of zero-sum games in economics.

\subsection{Method}
We take ViT as an example to provide an economic description of the model.

We name a group called the 	\textit{patch} clan, which consists of $n$ individuals. Each individual has a certain initial capital accumulation—either savings, debt, or no capital at all. However, the total assets of all individuals are fixed, denoted as \textit{output}. Later, the \textit{patch} clan admits an outsider named \textit{cls}, whose assets are 0. Currently, the clan has a total of $n+1$ members. The above describes the initial state.

The clan then holds $m$ meetings. Each meeting involves asset redistribution (with total assets remaining as \textit{output}). Asset transactions occur between every pair of the $n+1$ members. After the final (the $m$-th) transaction, \textit{cls} absorbs all savings and debts, and the capital of the $n$ original \textit{patch} clan members eventually becomes 0. (Strictly speaking, the initial state of the \textit{cls} token also accounts for a certain attribution contribution, but it is close to 0. For convenience of narration, we set it to 0 here; this setting does not affect the conclusion.)

The above process is essentially an economic phenomenon of a zero-sum game. Our task now is to evaluate the contribution of the $n$ individuals in the original \textit{patch} clan throughout this zero-sum game process. We use the cumulative asset holdings (CAH) of each \textit{patch} during the zero-sum game process to represent its contribution to this economic activity. \textbf{Specifically, we use IG\cite{cui2026neglectedbaselinemodelinterpretation} to evaluate asset holdings in the initial state and after each asset redistribution meeting, then accumulate the CAH of each \textit{patch} throughout the entire process as the final contribution. (\Cref{CAH})}

\begin{equation}\label{CAH}
  CAH_i = \sum_{l=0}^{L} IG^{(l)}_i
\end{equation}
where $i$ denotes the $i$-th patch token, $l$ denotes the $l$-th attention block, and $IG^{(l)}_i$ represents the IG score of the $i$-th patch token in the input features of the $l$-th attention block. Note that all contributions in the output features of the last attention block are attributed to the cls token, while the contributions of all patch tokens are zero. For this reason, we use input features for the formal formulation here.

The computation process of CAH itself embodies the core idea of IG. Fundamentally, IG characterizes the entire process, rather than focusing on only one single state as standard gradient-based methods do. CAH adheres to the same principle, as it captures every state along the path from the initial input state to the final state of the attention propagation process.
GAE evaluates the contribution of each individual attention unit via a gradient-based method, an approach that focuses solely on the current state. However, its real mechanism also embodies the core idea of IG. With the distinction that in GAE, only the scores of patch tokens within the cls token — derived from the accumulated result of first-order terms — exert a functional influence on the final output.

CAH is also an intuitive method for model evaluation, which represents a core task within model interpretability. The attention process is essentially a feature extraction process. For a well-performing model, the model rapidly extracts features from effective feature regions. In terms of CAH, this manifests as larger CAH score in effective feature regions and clearer boundaries with ineffective feature regions. 
However, it should be explicitly emphasized that this consistency-like evaluation pertains to model performance rather than CAH. The sole evaluation metric for CAH is the mean absolute value of IG attribution error (\Cref{cahexp}).

At this point, we have integrated both feature-based and process-based perspectives to complete the attribution interpretation of Transformers.

Note that the error of this cumulative asset holding method comes from the precision of IG, so there is no need for us to conduct experiments using flawed traditional evaluation metrics. The cumulative asset holding interpretation method can also assign weights to each asset state as needed.

\subsection{Experiments}\label{cahexp}
\begin{figure*}[]
  \centering
  \includegraphics[width=2\columnwidth]{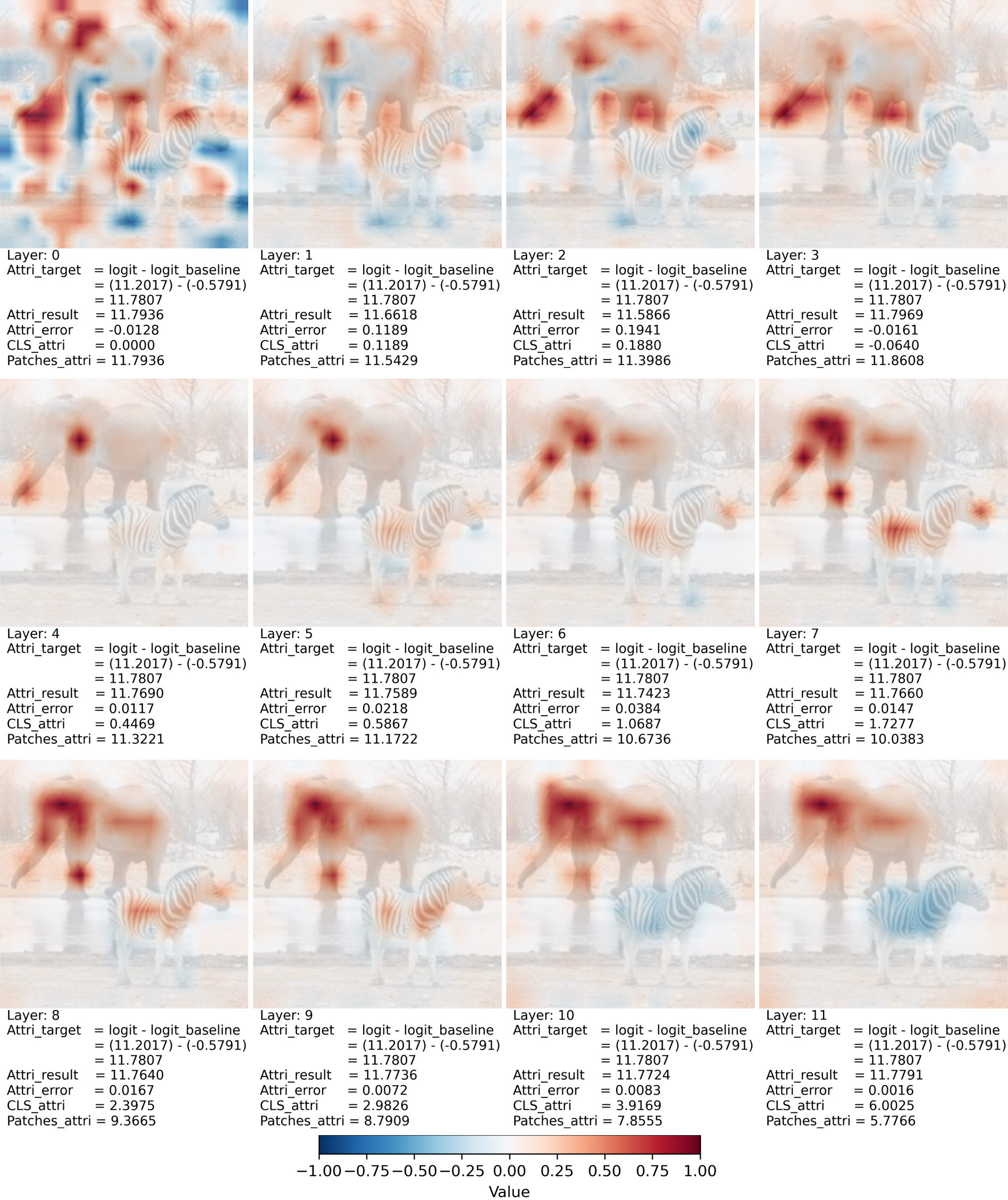}
  \caption{ViT\_base Integrated Gradients Interpretation\cite{cui2026neglectedbaselinemodelinterpretation} Based on Input Features of Different Attention Layers (Integration Steps: 256, Explanation Class: Tusker)}
  \label{7_layerig}
  \end{figure*}

\begin{figure*}[]
  \centering
  \includegraphics[width=2\columnwidth]{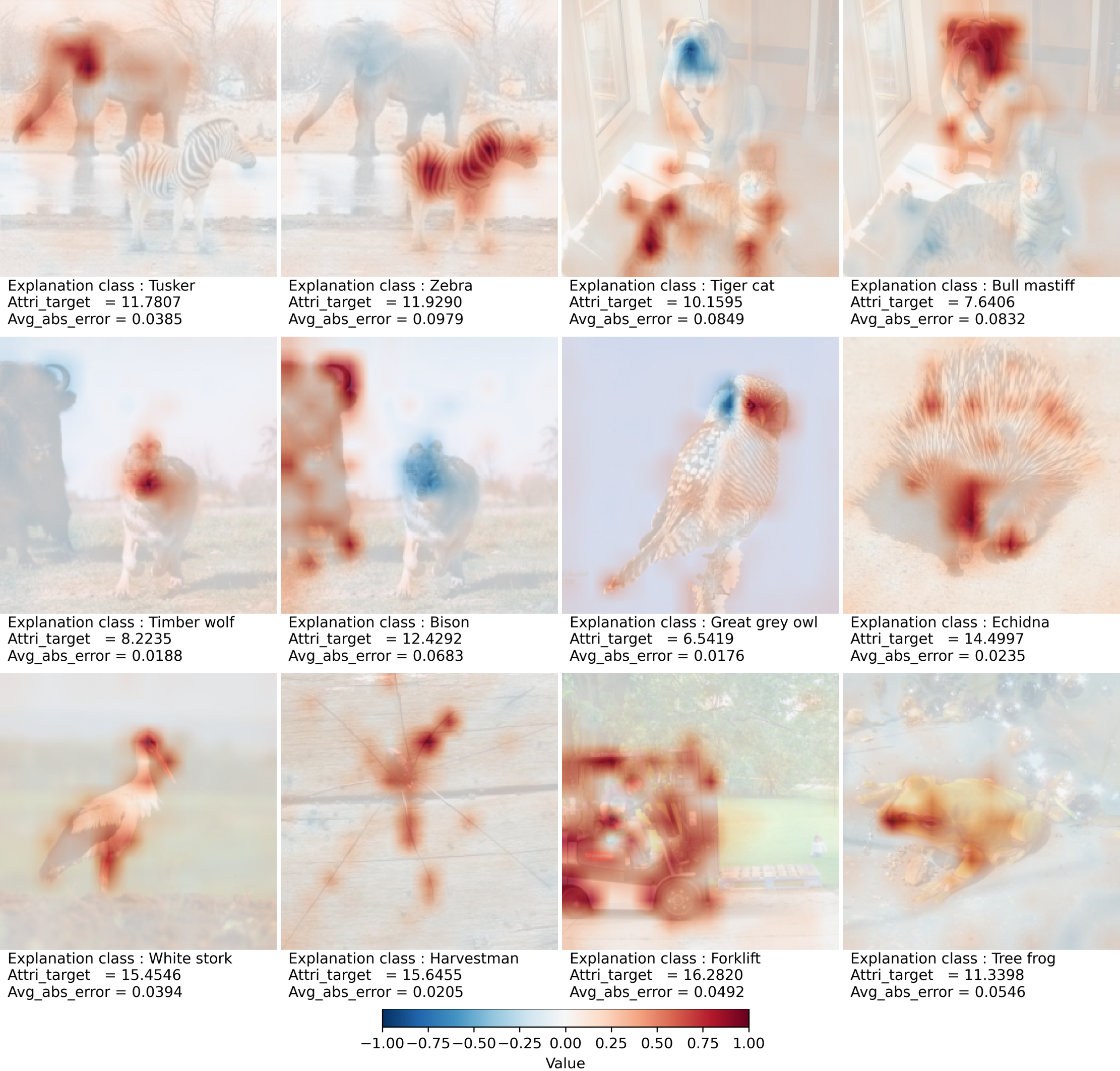}
  \caption{ViT\_base CAH Interpretation (Integration Steps: 256)}
  \label{7_leijizichan}
  \end{figure*}

As a foundational method, IG has been validated as a general-purpose analysis method across a wide range of models and datasets. Furthermore, gradient-based methods such as GAE are essentially coarse approximations of the IG method that incur larger attribution errors. By contrast, the CAH method proposed in this work is inherently an aggregation framework operating on IG outputs: it takes only the outputs of IG as input and involves no customized adaptations tailored to specific models, datasets, or tasks. Our proposed zero-sum game analogy remains valid even when the models only include patch tokens. We use models containing the cls token as the illustrative example  only because the application of IG suffers from limitations in models equipped with special tokens.
CAH enjoys a broader scope of application than IG. The generality of CAH represents a direct extension and expansion of the generality of IG. The experiments we present are not designed to prove the generality of CAH; instead, they aim to reveal the limitations of IG in models with special tokens and to demonstrate the final outputs of CAH. The error of CAH also originates exclusively from the attribution error of IG. The effectiveness of CAH is evaluated solely via the mean absolute value of IG attribution error. Accordingly, the experiments shown in this paper are provided solely for intuitive demonstration, rather than for verifying the generality and effectiveness of CAH.

\Cref{7_layerig} shows that in models with special tokens, IG can only assign part of the contribution to patch tokens, and cannot explain the contribution of patch tokens embedded within special tokens. \Cref{7_leijizichan} shows the contribution of patch tokens throughout the entire attention process measured by CAH. We express the error of CAH using the mean absolute value of IG attribution error per layer.

Note that the quality of an interpretation result should not be judged by whether it perfectly highlights feature regions, but by the magnitude of attribution error. Traditional consistency-based evaluation metrics are built upon the assumptions that the model is perfect and human cognition represents the optimal solution. Nevertheless, neither of these two assumptions generally holds. Attribution error is the basis on which we can trust the interpretation results.

\section{Conclusion}
\label{sec:discussion-zh}
This work stems from our advocacy for responsible XAI. We argue that model interpretation methods themselves must have sufficient interpretability; otherwise, there is no telling how many layers of X will need to be nested in front of XAI. Responsible model interpretation methods must themselves be sufficiently interpretable, clearly state the sources of error, and measure errors using accurate and reasonable evaluation methods.

In model training research, metrics are reliable and metric-oriented research is reasonable. For example, the original GLU paper \cite{GLU} candidly states: ``We offer no explanation as to why these architectures seem to work; we attribute their success, as all else, to divine benevolence." However, one of the most important tasks of XAI is to demystify this ``divine benevolence". XAI research can no longer rely on a metric-oriented research paradigm, especially since the currently adopted metrics are generally flawed.

Through model simplification, this paper studies GAE, a widely and profoundly influential representative model interpretation method, and reveals the real working mechanism and flaws of the technical pipeline it represents. \textbf{One major implication of this work is that, based on our findings, all GAE-related applications and subsequent methods adopting similar technical pipelines may need to be re-examined.} We argue that GAE is an interpretation method focusing on the attention process. After pointing out the working mechanism and flaws of GAE, we propose CAH, a more reasonable process-based Transformer interpretation method from the perspective of economic zero-sum games. This method, like GAE, focuses on the entire attention process. Meanwhile, it applies IG to solve the precision problem of gradient-based methods such as GAE. More importantly, it addresses the difficulty of fully attributing model outputs to patch tokens when IG is directly applied to models with special tokens. The model simplification research method and the analysis of additive operations adopted in this study may provide inspiration for other research works in this field.

\FloatBarrier



\bibliographystyle{IEEEtran}
\bibliography{aaai2026}

\end{document}